\documentclass[nohyperref,12pt]{iopart}
\usepackage[english]{babel}
\usepackage{amssymb,graphicx,enumerate,xcolor,latexsym,theorem}
\usepackage[numbers]{natbib}
\usepackage{iopams}

\renewcommand\harvardurl[1]{\textbf{URL:} \url{#1}}
\usepackage[pagebackref=true,breaklinks=true,letterpaper=true,colorlinks,bookmarks=false]{hyperref}

\expandafter\let\csname equation*\endcsname\relax
\expandafter\let\csname endequation*\endcsname\relax
\usepackage{amsmath}
\usepackage{cleveref}

\usepackage{pgfplots}
\DeclareUnicodeCharacter{2212}{−}
\usepgfplotslibrary{groupplots,dateplot}
\usetikzlibrary{patterns,shapes.arrows}
\pgfplotsset{compat=newest}

\newcommand{\nobracket}{}

\newcommand{\tmop}[1]{\ensuremath{\operatorname{#1}}}

{\theorembodyfont{\rmfamily}}
\usepackage[small]{caption}

\usepackage{tikz}
\usepackage{tikzscale}
\usepackage{layouts}
\newlength\tikzwidth
\newlength\tikzheight

\begin{document}
\bibliographystyle{unsrtnat}

\title{
  Looking at the posterior: accuracy and uncertainty of neural-network predictions
}

\author{Hampus Linander$^{1,2}$\footnote{email: linander@chalmers.se}, Oleksandr Balabanov$^{3}$, Henry Yang$^{1}$, Bernhard Mehlig$^{1}$}

\address{$^{1}$ Department of Physics, University of Gothenburg, 41296 Gothenburg, Sweden}
\address{$^{2}$ Department of Mathematical Sciences, Chalmers University of Technology, University of Gothenburg, 41296 Gothenburg, Sweden}
\address{$^{3}$ Department of Physics, Stockholm University, 10691 Stockholm, Sweden}

\begin{abstract}
  Bayesian inference can quantify uncertainty in the predictions of neural networks using posterior distributions for model parameters and network output. By looking at these posterior distributions, one can separate the origin of uncertainty into aleatoric and epistemic contributions. One goal of uncertainty quantification is to inform on prediction accuracy.
  Here we show that prediction accuracy depends on both epistemic and aleatoric uncertainty in an intricate fashion that cannot be understood in terms of marginalized uncertainty distributions alone. How the accuracy relates to epistemic and aleatoric uncertainties depends not only on the model architecture, but also on the properties of the dataset.
  We discuss the significance of these results for active learning and introduce a novel acquisition function that outperforms common uncertainty-based methods. To arrive at our results, we approximated the posteriors using deep ensembles, for fully-connected, convolutional and attention-based neural networks.

\end{abstract}

\maketitle

\section{Introduction}
The user of an artificial neural-network wants to know when the prediction of the model is accurate and trustworthy. When target ground truth is unavailable, as is usually the case, one must instead rely upon surrogate measures that correlate with accuracy and trustworthiness in a robust way. Uncertainty quantification aims to provide such measures. Recently there has been an intensive effort towards a better understanding of uncertainty of neural-network predictions \citep{gawlikowski2022survey,ABDAR2021243}.
To quantify this uncertainty in a way that informs on the efficacy of the model, and to identify its sources, is of key significance in many applications of machine-learning algorithms using neural networks, from real-time predictions to active learning {\citep{gustafsson2020evaluating,al1,lee2017deep,mervin2021uncertainty,lye2020deep,liu2020probabilistic,butler2021interpretable}}.

When the outputs of neural networks can be viewed as probability distributions over possible output values, certain distributional measures naturally capture the uncertainty of the network predictions. For instance, if the output distribution is sharply peaked, one might expect the prediction to be accurate. To which extent
this expectation is borne out, depends not only on the model architecture and parameters, but also on the input data (for example whether it is from a domain the model has knowledge about). 

 Bayesian inference {\citep{Hinton1995BayesianLF}} provides a theoretical framework to reason about the conditional distribution of model parameters, and of the model output, given the available training data. More precisely, given a neural network with parameters $\theta$, a prior $p(\theta)$, and a training dataset $\mathcal{D} = \left\{(x_{1},y_{1}), (x_{2}, y_{2}),\cdots\right\}$ of pairs $(input, target)$,
Bayesian arguments determine a distribution over the neural-network parameters $p(\theta | \mathcal{D})$ \citep{mackay}. This so-called posterior distribution tells us
the probability of different model parameters given the training dataset. Using
this posterior distribution for the parameters, a corresponding posterior distribution of the neural-network predictions,
\begin{align}
  p (y | \mathcal{D}, x \nobracket) & = \int_{\theta} p (y | \theta, x
  \nobracket) p (\theta | \mathcal{D} \nobracket) d \theta, \label{eq:bayesian_mean}
\end{align}
can be derived.
The posterior predictive distribution in Eq.~\eqref{eq:bayesian_mean} is the
 marginalization over model parameters $\theta$, conditioned on a particular input $x$
 that is either previously unseen or contained in the training dataset. Together, the
 posterior distribution of model parameters and the posterior predictive distribution
 characterize the knowledge of the model, and their entropies provide measures of
 uncertainty.

The entropy of the posterior parameter distribution measures the uncertainty of the
model parameters, and as such is \textit{epistemic}. Uncertainty stemming from the input
data is often referred to as \textit{aleatoric}, and it is related to the entropy of
the posterior predictive distribution {\citep{KIUREGHIAN2009105}}. Bayesian modeling
makes the distinction between epistemic and aleatoric uncertainty clear. The posterior
probability $p (\theta | \mathcal{D} \nobracket)$ of the neural-network parameters $
\theta$ given the observations $\mathcal{D} $ determines the epistemic uncertainty.
The aleatoric part, by contrast, is determined by the likelihood $p (y | \theta, x
\nobracket)$ given model parameters $\theta$ and input data $x$. The entropy of the
posterior predictive distribution in Eq.~\eqref{eq:bayesian_mean} contains through the
two factors in the integrand a mixture of aleatoric and epistemic uncertainty, referred
to as predictive uncertainty.

Hence, the decomposition of uncertainty into aleatoric and epistemic parts
can be quantified by the entropy of the posterior predictive distribution in
Eq.~\eqref{eq:bayesian_mean}, containing both aleatoric and epistemic contributions,
and the posterior parameter distribution $p(\theta | \mathcal{D})$. 
The epistemic uncertainty associated with a particular input $x$ can then be quantified
by the conditional mutual information $I(\theta; y|x, \mathcal{D}) = H(\theta| \mathcal{D}) - E_{p(y|x,\mathcal{D})}[H(\theta|(x,y), \mathcal{D})]$, where $H(\cdot|\cdot)$ denotes conditional entropy,
between parameters and predictive distribution conditioned on the input $x$ and training
dataset $\mathcal{D}$ \cite{mackay}.
However, it
turns out that this separation of uncertainty into aleatoric and epistemic parts
can be hard to use in practice, as evident by the following two examples. The
first one comes from active learning: It is found that using a decomposition of
predictive uncertainty into aleatoric and epistemic parts does not necessarily
improve sample selection \citep{beluch2018power, nguyen2019epistemic, gal2017deep, chitta2018large}. 
This is surprising, since high epistemic uncertainty is a
natural criterion for samples where the model can hope to learn something new.
The second example relates to safety critical applications such as medical
diagnosis. For medical image analysis in particular, uncertainty quantification
has been used to improve model precision, and to guide clinical assessment
\citep{ruhe2019bayesian,hu2019supervised,reinhold2020validating,ghoshal2,ghoshal}.
By selecting for low epistemic and aleatoric uncertainty one could hope to increase
prediction accuracy at the cost of recall. Again, marginalized measures in terms of
epistemic and aleatoric parts do not seem to provide better precision over predictive
uncertainty \cite{nair2020exploring}. Why is it hard to use the decomposition into
aleatoric and epistemic uncertainty effectively? To answer this question we evaluate the
correlation between accuracy and uncertainty quantification using the joint distribution
of predictive uncertainty and epistemic uncertainty, as measured by the entropies of the
posterior predictive distribution and the posterior parameter distribution.

Our results show that the accuracy of a model has non-trivial correlations with
the combination of predictive and epistemic uncertainty, and that this correlation
depends on model architecture and dataset. We test these insights in application by proposing and evaluating a novel acquisition function based on expected accuracy. 
Using the joint distribution of predictive
uncertainty and epistemic uncertainty, we quantify how the approximate posteriors of
three common neural-network architectures for image classification differ from each
other and how they depend on data-distributional shifts in the form of impulse noise
\citep{AWAD2019746} for MNIST \citep{mnist} and CIFAR \citep{krizhevsky2009learning}.
We conclude that the joint distribution contains important information regarding
model accuracy, and that it needs to be calibrated for a particular model architecture
and dataset. 

\section{Contributions}

\begin{itemize}
\item We quantify the joint distribution of prediction accuracy, predictive uncertainty
and epistemic uncertainty for the MNIST \cite{mnist} and CIFAR \cite{cifar10} datasets.
The non-trivial patterns we find makes it clear that model accuracy cannot be understood
in terms of marginalized uncertainty measures.
  \item We introduce a novel acquisition function for active learning that outperforms acquisition using marginalized uncertainty distributions.
  \item We use the joint distribution of predictive entropy and conditional mutual
information between parameters and targets to quantify the variability of uncertainty
measures over different model architectures and data-distributional shifts.
  \item We show that neural networks with different architectures can disagree about the origin of uncertainty for data-distributional shifts in image classification tasks.
\end{itemize}
\section{Related work}
 Ref. \citep{pmlr-v80-depeweg18a} uses aleatoric and epistemic uncertainty separately for active learning on low dimensional regression and classification tasks, concluding that epistemic uncertainty provides a better criterion for selecting training samples than the predictive uncertainty. This provides a clear example of where the intuition of epistemic uncertainty being a good sample selection criteria holds. In contrast, this does not extend to higher dimensional computer vision domains where epistemic uncertainty is no longer superior \cite{chitta2018large}. In the context of reinforcement learning, Ref. \citep{pmlr-v80-depeweg18a} uses a combination of aleatoric and epistemic uncertainty to find balanced policies.  The joint distribution, architecture dependence, and dataset dependence is not considered. In Ref. \citep{ghoshal}, the correlation between accuracy and uncertainty is quantified for image classification. Correlation with accuracy is shown for predictive uncertainty and epistemic uncertainty separately, the architecture and dataset dependence is not discussed. The joint distribution of aleatoric and epistemic uncertainty is considered in Ref. \citep{ghoshal2}, where the correlation between predictive probabilities and the joint uncertainty distribution is quantified in the context of medical image semantic segmentation. Using a fixed residual U-Net architecture and  datasets for semantic segmentation it was shown that there is a correlation between predictive probabilities and the uncertainty measures, and that accuracy for their semantic segmentation model on the considered datasets shows correlation with epistemic uncertainty. In terms of open questions, Ref. \citep{ghoshal2} highlight the effect of data-distributional shifts, model architecture, and data modality on the quality of uncertainty quantification. These are in line with two of our target questions on how the perceived origin of uncertainty depends on model architecture and how the dataset impacts uncertainty quantification. Quantifying uncertainty under data-distributional shifts was investigated in Ref. \citep{googleuncertainty} where accuracy, calibration and entropy of the posterior predictive is evaluated for different shifts introduced in Ref. \citep{cifar10c}. Here, the epistemic uncertainty is not considered and the model dependence of the uncertainty measures is not analyzed.

\section{Background}
\subsection{Bayesian inference}
An artificial neural network $f$ with parameters $\theta \in \mathbb{R}$ can be seen as a map from input space $X$ to the output space $Y$ \citep{bm}, where we take $Y$ to be the space of distributions over possible outcomes so that the likelihood is given by $p(y| x, \theta) = f(y, x ; \theta)$.
Bayesian inference {\citep{sivia2006data}} allows us to reason about
uncertainty in terms of posterior distributions for the parameters of a model.
With a training dataset $\mathcal{D} \in X \times Y$ corresponding to observations $(x_t, y_t)$ with $t\in\{1\cdots N\}$, the posterior distribution for the neural-network parameters $p(\theta | \nobracket
\mathcal{D})$ can be calculated using Bayes theorem in terms of the likelihood as
\begin{equation}
  p(\theta | \mathcal{D}) = \frac{p(\mathcal{D} | \theta)p(\theta)}{p(\mathcal{D})},
\end{equation}
where $p(\mathcal{D} | \theta)$ (assuming independent samples in the dataset $\mathcal{D}$) can be expressed as $p(\mathcal{D}|\theta) = \prod_{i=1}^{N} f(y_{i}, x_{i}; \theta)$, $p(\theta)$ the parameter prior and the evidence $p(\mathcal{D})$ is the marginalization over the parameter prior.

A prediction, or more generally a distribution over possible outputs, is computed using the posterior predictive distribution in Eq.~\eqref{eq:bayesian_mean} by
marginalizing over the parameters.
Each parameter configuration is weighted by its posterior probability given the training
data. For large neural-network architectures, it is in general very challenging to evaluate the high-dimensional integral over $\theta$ \cite{izmailov2021bayesian}.
To implement Bayesian inference for neural networks, the
posterior needs to be approximated.

There are a number of different methods
available that range from computationally expensive Monte-Carlo simulations
{\citep{Hinton1995BayesianLF}} to more efficient dropout approximations
{\citep{1703.04977,balabanov2023bayesian}}, as well as simpler ensembling methods
{\citep{lakshminarayanan2017simple}}. The most accurate approximations
to the true posterior rely on the Hamiltonian Monte-Carlo (HMC) methods
\citep{Duane1987,Mehlig1992,Neal1996}. Through intensive computational efforts, HMC
methods have recently been used to approximate the posteriors of larger convolutional
neural networks such as a 20-layer ResNet \citep{izmailov2021bayesian}. The HMC
computations show that simpler approximation schemes such as ensembling and variational
inference can fail to accurately describe the true posterior, but that ensembles often
provide more accurate posteriors than more advanced methods. Here we use deep ensembles, as
described in \ref{sec:ensemble}, to approximate the Bayesian posteriors of the neural networks \cite{hoffmann2021deep}.

\subsection{Uncertainty quantification}
\label{sec:uncertainty_quantification}
For classification over a discrete set of $M$ classes, the entropy {\citep{shannon}} of the predictive distribution
\begin{equation}
  H(p) = -\sum_{c=1}^{M}p( {c} )\log (p( {c} )), \label{eq:entropy}
\end{equation}
provides a measure of the information content and thus its uncertainty.

In terms of the ensemble approximations of the posterior predictive distribution, see appendix Eq.~\eqref{eq:ppd} for details, this takes the form
\begin{equation}
  H\big(p(y| x, \mathcal{D})\big) = -\sum_{c=1}^{M}\left(\frac{1}{N}\sum_{ i=1 }^N f_{c}(x; \theta^{(i)})\right)\log \left(\frac{1}{N}\sum_{ i=1 }^N f_{c}(x; \theta^{(i)})\right), \label{eq:posterior_entropy_approx}
\end{equation}
where $H(p(y|x, \mathcal{D}))$ is the entropy of
the posterior predictive distribution, referred to here as predictive uncertainty.
The entropy of $p(y|x, \mathcal{D})$ in Eq.~\eqref{eq:posterior_entropy_approx}, as a measure of uncertainty, contains contributions that are both epistemic and aleatoric.

Epistemic uncertainty stems from uncertainty in model parameters. This is captured by
the shape of the posterior distribution of these parameters. To quantify the epistemic
uncertainty associated with a single data sample $x$, we can ask how this shape changes
when the dataset is extended to include $x$~\citep{mackay}.
The epistemic uncertainty associated with a data sample
$x$ can be quantified by the expected change in entropy of the model parameter posterior distribution $p(\theta|\mathcal{D})$ when $x$ is added to the observations~\citep{mackay},
\begin{align}
  I(x) & = H (p(\theta | \mathcal{D}) \nobracket) - E_{p(y | x \nobracket, \mathcal{D})} \left[H \big(p(\theta | \nobracket \mathcal{D}, (x, y))\big)\right]\; \label{eq:delta_H_theta}.
\end{align}
The entropy difference in Eq.~\eqref{eq:delta_H_theta} is the conditional mutual information $I(\theta; y | x, \mathcal{D})$ between the parameter posterior and the predictive posterior distribution for the new sample point $(x,y)$. Since mutual information is symmetric, this can be written in terms of the entropy of the predictive distribution instead:
\begin{align}
  I(x) & =  H (p(y | x, \mathcal{D}) \nobracket) - E_{p
  (\theta | \mathcal{D} \nobracket)} \left[H \big(p(y | \theta, x) \nobracket\big)\right]\;,
  \label{deltaH}
\end{align}
where $H (p(y | \theta, x \nobracket))$ is the entropy of the likelihood $p(y | \theta, x)$.
See \ref{sec:toy_model} for an example of how this relation manifests in a toy model.
The first term in Eq.~\eqref{deltaH} is the entropy of the posterior predictive distribution given the dataset $\mathcal{D}$,
whereas the second term is the expected value of the likelihood entropy over the model posterior distribution.

Using the approximate posteriors from the ensemble method, the first term in Eq.~\eqref{deltaH} is given by Eq.~\eqref{eq:posterior_entropy_approx} and the second term is given by
\begin{equation}
E_{p(\theta | \mathcal{D} \nobracket)} \left[H \big(p(y | \theta, x) \nobracket\big)\right] = \frac{1}{N} \sum_{i=1}^{N}\sum_{c=1}^{M}f_{c}(x; \theta^{(i)})\log \left(f_{c}(x; \theta^{(i)})\right).\label{eq:expected_H}
\end{equation}
Together, Eq.~\eqref{eq:posterior_entropy_approx} and \eqref{eq:expected_H} provide a concrete way to evaluate epistemic uncertainty as defined by Eq.~\eqref{deltaH} in practice.

The epistemic uncertainty in Eq.~\eqref{deltaH} is large when the posterior predictive entropy is large and the mean likelihood entropy is small. In terms of the ensemble members this corresponds to the situation where each member has a sharp distribution but they disagree about the mean. Large aleatoric uncertainty is ascribed to broad output distributions from the individual members $f(\cdot; \theta^{(i)})$, also implying a large posterior predictive entropy.
Since entropy is positive, both the term $H(p(y|x, \mathcal{D}))$ and the expected entropy of the likelihood in Eq.~\eqref{deltaH} are positive, and so the entropy difference in Eq.~\eqref{deltaH} is bounded by the posterior entropy in Eq.~\eqref{eq:posterior_entropy_approx}, resulting in the inequality
\begin{equation}
  I(x) \leq H(x)\;, \label{eq:bound}
\end{equation}
i.e.~a large epistemic uncertainty implies a large posterior predictive entropy. This can be seen by noting that a collection of sharp member distributions that disagree about the mean necessarily adds up to a broad mean distribution. One way to describe a sample with large epistemic uncertainty is that the model might fit the data well, but in many different ways. The parameters with high posterior probability could all result in sharp distributions, whereas the full posterior can be broad {\citep{batchbald}}.

Since the epistemic uncertainty in Eq.~\eqref{deltaH} measures the change in posterior entropy, a model can have irrelevant parameters that contribute to a broad posterior distribution, but the change in posterior entropy can still be small when adding a given sample to the training set. If the parameter posterior for some irrelevant parameter is equally broad after we add sample $x$, then we do not want to consider this as a point of high epistemic uncertainty.

Even though entropic measures are a natural starting point for quantifying uncertainty of predictions, they have been shown to possess a number of unwanted properties such as ataining maximal values in un-intuitive scenarios \cite{wimmer2023quantifying}.
In this work we take a pragmatic view and consider uncertainty quantification based on entropy as a flawed, but still practically useful tool to understand neural network predictions.

\section{Methods}
\subsection{Datasets}
\label{sec:datasets}
Two datasets are used for all numerical experiments, MNIST {\citep{mnist}} and a
grayscale version of CIFAR10 {\citep{cifar10}}. We choose these simple datasets to be able to train ensembles efficiently and still be in the domain of realistic images with CIFAR10. To be able to compare relative shifts in uncertainty measures between the two datasets we choose to work with a grayscale version of CIFAR denoted CIFAR10G so that the data-distributional shift acts in exactly the same way for both MNIST and CIFAR10G.
MNIST is an example of an image classification dataset with
minimal complexity, whereas CIFAR10 provides a more realistic data distribution for image classification.
The grayscale conversion for the RGB data from CIFAR10 is given by the standard BT.601 luminance $Y = 0.2989 r + 0.5870 g + 0.1140 b$.
We apply impulse noise \cite{AWAD2019746}, a common corruption present in digital images, to the original datasets (MNIST, CIFAR10G) where the strength of the perturbation is controlled by a noise parameter $\alpha$.
For ``salted'' noise with parameter $\alpha > 0$ a random sample of pixels of size $\alpha N_{\tmop{pixels}}$ are given the maximum value $1.0$, and for ``peppered'' noise with $\alpha < 0$ a
corresponding amount of pixels are set to $0$.
We pick  $N_{\alpha}$ distinct values between $\alpha_{\text{min}}=-0.3$ and $\alpha_{\text{max}}=0.3$.

\begin{figure}[ht]
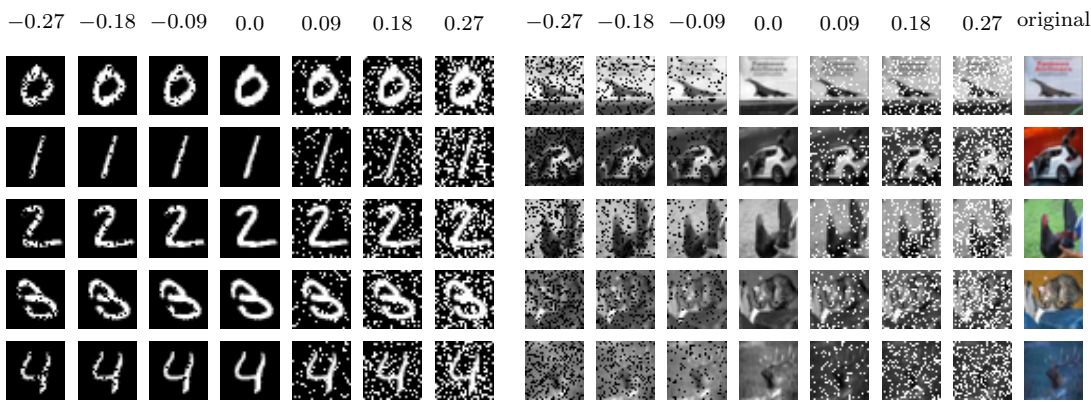

  \centering
  \setlength{\tikzwidth}{67pt}
  \setlength{\tikzheight}{67pt}
  \input{figures/mnist_salt_and_peppered.tikz}
  \input{figures/cifar_gray_salt_and_peppered.tikz}
  \caption{MNIST (left) and CIFAR10G (right) with impulse noise parameterized by
  $\alpha$. Darker colors indicate a value closer to $0$. For CIFAR10G the original RGB image is shown in the right-most column. The CIFAR10 classes in the examples are, from top to bottom: airplane, automobile, bird, cat, deer.  \label{fig:mnist_salt}}
\end{figure}

MNIST consists of 60000 grayscale images with resolution $28{\times}28$. The original CIFAR10 dataset consists of 60000 RGB images with resolution $32{\times}32$ that we resample to a single grayscale channel.
See figure \ref{fig:mnist_salt} for examples of the different noise levels.

\subsection{Network architectures}
\label{sec:models}
\begin{table}
  \centering
  \begin{tabular}{ccll}
    Model & Parameters & MNIST & CIFAR10G \\
    \hline
    Dense & 118k & $99.5\% \pm 0.07\%$& $45.8\% \pm 0.5\%$\\
    CNN & 836k & $99.9\% \pm 0.06$ & $74.1\% \pm 0.95\%$\\
    Swin & 147k & $97.3\% \pm 0.1\%$ & $66.2\% \pm 1.3\%$\\
  \end{tabular}
  \caption{Summary of the three different neural-network architectures used for comparisons. The accuracy for MNIST and CIFAR10G is over the validation datasets with standard deviation calculated for 10 separate trainings. The dense model uses 3 layers (128, 128, 10) with ReLU activations, CNN is model A in \cite{cnn} and Swin is a shifted window transformer\cite{swin} based on the KERAS vision implementation\cite{Dagli2021}.}
  \label{tab:architectures}
\end{table}
In order to quantify how the architecture affects the uncertainty estimates, we use three neural-network architectures: fully connected (Dense), convolutional (CNN) and attention-based (Swin) neural networks.
The fully connected neural network has a three-layer architecture with two 128-neuron hidden layers using ReLU activations.
The convolutional neural network is identical to model A in~\cite{cnn}, a simple fully convolutional model with 5 layers using max-pooling for spatial down-sampling and ReLU activations.
Finally, we also use a small version of a shifted windows transformer (Swin)~\cite{swin}, a popular attention-based model for computer vision. All architectures use a softmax output layer.
See Table \ref{tab:architectures} for a summary of the model sizes and baseline accuracy on the target datasets.

\subsection{Active learning}
\label{sec:method_AL}
One way of improving the accuracy of a neural-network model is to extend the training set. Active learning \citep{angluin1988queries,atlas1989training,liu2022survey,settles2012active} makes use of the network predictions to choose inputs that contribute the most to increased accuracy. 
Which inputs to choose can be phrased in terms of a so-called acquisition function \citep{martinez2014bayesopt}, that determines how samples are picked from the pool of unlabelled data.
Common acquisition functions include BALD scoring \cite{bald} that picks samples with highest epistemic uncertainty
as measured by mutual information $I(x)$ in Eq.~\eqref{deltaH}. Another common acquisition function is or to choose
samples with highest predictive uncertainty \cite{lewis1995sequential} quantified by $H(y|x, \mathcal{D})$ in
Eq.~\eqref{eq:posterior_entropy_approx}. We refer to these two methods as BALD and max entropy respectively.
Based on our results for the correlation between accuracy and the joint distribution $(H, I)$ we propose a novel acquisition function that picks samples with lowest expected accuracy in section \ref{sec:results_al}.

\section{Results}
\label{sec:results}
\subsection{Accuracy}
\label{sec:accuracy}

\begin{figure}[t!]
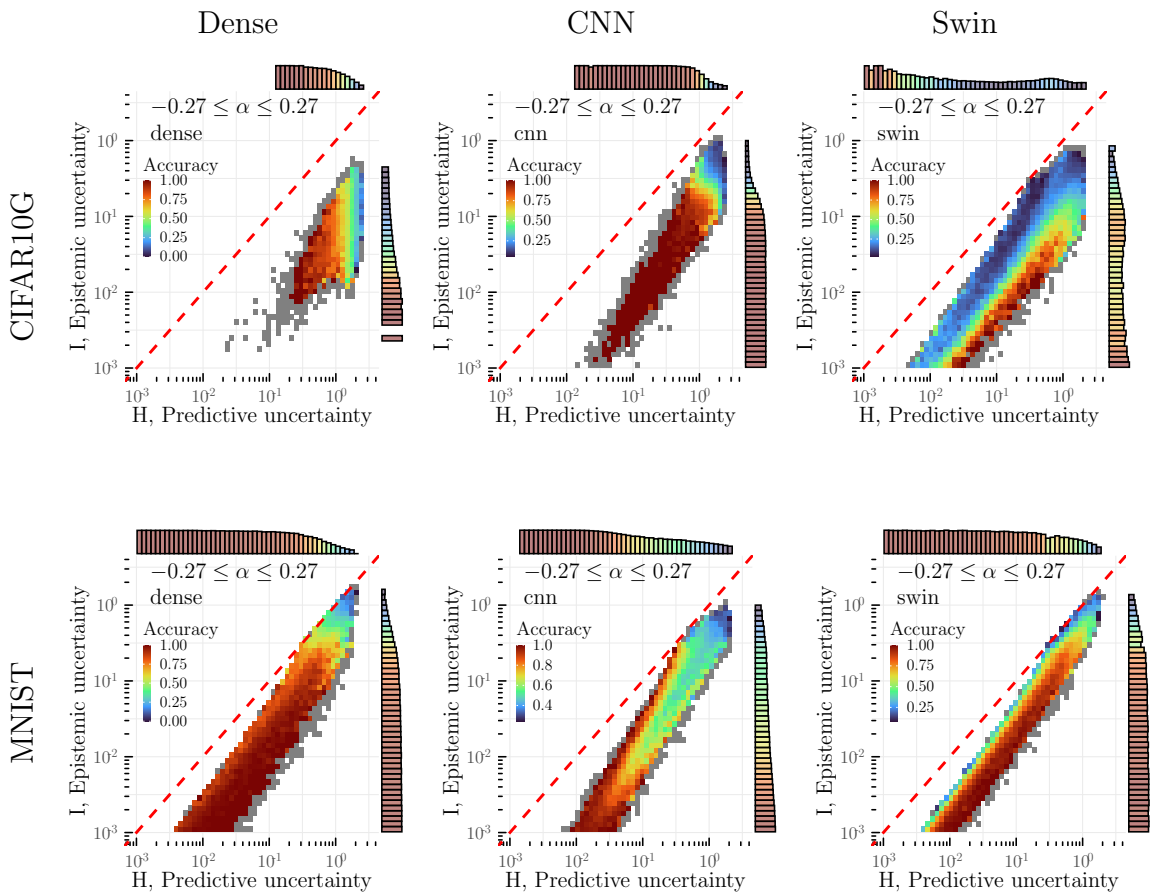

  \centering
  \begin{minipage}{\textwidth}
  \raisebox{5em}{\rotatebox{90}{CIFAR10G}}
  \begin{minipage}[b]{0.33\textwidth}
  \centering 
  Dense \\
  \input{ep/ensemble_cifar_dense.tex}
  \end{minipage}
  \hspace{-1.5em}
  \begin{minipage}[b]{0.33\textwidth}
  \centering 
  CNN \\
  \input{ep/ensemble_cifar_cnn.tex}
  \end{minipage}
  \hspace{-1.5em}
  \begin{minipage}[b]{0.33\textwidth}
  \centering 
  Swin \\
  \input{ep/ensemble_cifar_swin.tex}
  \end{minipage}
  \end{minipage}
\begin{minipage}[t]{\textwidth}
\raisebox{5em}{\rotatebox{90}{MNIST}}
  \input{ep/ensemble_dense.tex}
  \hspace{-1.5em}
  \input{ep/ensemble_cnn.tex}
  \hspace{-1.5em}
  \input{ep/ensemble_swin.tex}
  \end{minipage}
  
\caption{\label{fig:cifar_acc}\label{fig:mnist_acc}Accuracy distribution in the
$(H, I)$-plane, and marginal accuracy distributions (above and to the right of each panel), for a union over noise levels
$-0.27 \leq \alpha \leq 0.27$ evaluated on CIFAR10G (top row) and MNIST (bottom row). Grey bins indicate less than 10 inputs in the corresponding region.
Prediction accuracy conditioned on the joint distribution of posterior predictive
entropy $H$ and epistemic uncertainty $I$ is shown for the ensemble posterior. }
\end{figure}

Figure~\ref{fig:cifar_acc} shows the accuracy of predictions of three neural networks for MNIST and CIFAR10G as a function of predictive and epistemic uncertainty, evaluated on the union of all noise levels including the uncorrupted test set.

For CIFAR10G in Figure~\ref{fig:cifar_acc}, the simple fully connected neural network in the first panel has a pronounced correlation between the predictive uncertainty and model accuracy. In the center panel, the convolutional model shows a different pattern where both epistemic and predictive uncertainty correlates equally with accuracy. For the attention-based model in the last panel, there is a more intricate relation. Decreasing epistemic uncertainty correlates with higher accuracy, but predictive uncertainty for a fixed epistemic value does not.
In general, lower epistemic and aleatoric uncertainty does not always imply higher accuracy.

Turning to MNIST in Figure~\ref{fig:mnist_acc}, the first panel shows that the correlation between accuracy and predictive uncertainty is not as pronounced as in the first panel in Figure~\ref{fig:cifar_acc}. In the domain of MNIST where the dense model has higher accuracy there is instead a stronger correlation between decreasing epistemic uncertainty and prediction accurac. The center panel shows that for the convolutional neural network, for a fixed moderate predictive uncertainty, the accuracy does not monotonically increase with epistemic uncertainty. On the other hand, the attention-based Swin architecture in the right panel shows a monotonically increasing accuracy with decreasing epistemic uncertainty with similar structure as for CIFAR10G.

The moments of the joint uncertainty distribution for CIFAR10G at a fixed distributional shift are summarized in Table~\ref{tab:method_cifar} together with average prediction accuracies. For MNIST, the corresponding joint uncertainty distribution at fixed data-distributional shift is given in Table~\ref{tab:method_mnist}. For reference, the moments and accuracy on the unshifted test set is shown in Table~\ref{tab:method_cifar_validation}. Comparing the average accuracy over the shifted dataset in Table~\ref{tab:method_cifar} with the accuracy on the validation set in Table~\ref{tab:method_cifar_validation}, we see that the dense model retains its accuracy better than the more complex architectures. For each individual architecture, a decrease in average predictive and epistemic uncertainty is accompanied by an increase in average accuracy.

\begin{table}[t!]
  \centering
  \begin{tabular}{|l||lll|}
    \hline
    Architecture & Acc. & \multicolumn{1}{c}{H} & \multicolumn{1}{c|}{I} \\
    \hline
    Dense & $41\%$ & $1.66 \pm 0.39$ & $0.10 \pm 0.04$ \\
\hline
CNN & $20\%$ & $1.92 \pm 0.22$ & $0.39 \pm 0.11$ \\
\hline
Swin & $20\%$ & $1.12 \pm 0.47$ & $0.27 \pm 0.14$ \\
    \hline
  \end{tabular}
  \caption{Mean and standard deviation for predictive and epistemic uncertainty for fully connected (Dense), convolutional (CNN) and attention-based (Swin) models with posterior approximations using ensembles evaluated on CIFAR10G at fixed data-distributional shift $\alpha=0.09$. \label{tab:method_cifar}}
\end{table}

\begin{table}[t!]
  \centering
  \begin{tabular}{|l||lll|}
    \hline
    Architecture & Accuracy & \multicolumn{1}{c}{H} & \multicolumn{1}{c|}{I} \\
    \hline
    Dense & $65\%$ & $0.54 \pm 0.47$ & $0.35 \pm 0.32$ \\
\hline
CNN & $73\%$ & $0.67 \pm 0.54$ & $0.25 \pm 0.22$ \\
\hline
Swin & $58\%$ & $0.69 \pm 0.44$ & $0.44 \pm 0.30$ \\
    \hline
  \end{tabular}
  \caption{Mean and standard deviation for predictive and epistemic uncertainty for fully connected (Dense), convolutional (CNN) and attention-based (Swin) models with posterior approximations using ensembles for MNIST at fixed data-distributional shift $\alpha=0.09$. \label{tab:method_mnist}}
\end{table}

\begin{table}[t!]
  \centering
  \begin{tabular}{|l||lll|}
    \hline
    Architecture & Acc. & \multicolumn{1}{c}{H} & \multicolumn{1}{c|}{I} \\
    \hline
    Dense & $45\%$ & $1.58 \pm 0.41$ & $0.07 \pm 0.04$ \\
\hline
CNN & $75\%$ & $1.07 \pm 0.64$ & $0.14 \pm 0.10$ \\
\hline
Swin & $65\%$ & $0.90 \pm 0.55$ & $0.08 \pm 0.06$ \\
    \hline
  \end{tabular}
  \caption{Mean and standard deviation for predictive and epistemic uncertainty and accuracy for fully connected (Dense), convolutional (CNN) and attention-based (Swin) models with posterior approximations using deep ensembles evaluated on CIFAR10G without data-distributional shift. \label{tab:method_cifar_validation}}
\end{table}

\subsection{Active learning}
\label{sec:results_al}
\begin{figure}[t]
\centering
\begin{tikzpicture}[x=1pt,y=1pt]
\definecolor{fillColor}{RGB}{255,255,255}
\path[use as bounding box,fill=fillColor,fill opacity=0.00] (0,0) rectangle (200.66,150.49);
\begin{scope}
\path[clip] ( 21.74, 19.53) rectangle (126.36,146.99);
\definecolor{drawColor}{gray}{0.92}

\path[draw=drawColor,line width= 0.2pt,line join=round] ( 21.74, 22.78) --
	(126.36, 22.78);

\path[draw=drawColor,line width= 0.2pt,line join=round] ( 21.74, 54.58) --
	(126.36, 54.58);

\path[draw=drawColor,line width= 0.2pt,line join=round] ( 21.74, 86.39) --
	(126.36, 86.39);

\path[draw=drawColor,line width= 0.2pt,line join=round] ( 21.74,118.20) --
	(126.36,118.20);

\path[draw=drawColor,line width= 0.2pt,line join=round] ( 29.44, 19.53) --
	( 29.44,146.99);

\path[draw=drawColor,line width= 0.2pt,line join=round] ( 55.77, 19.53) --
	( 55.77,146.99);

\path[draw=drawColor,line width= 0.2pt,line join=round] ( 82.10, 19.53) --
	( 82.10,146.99);

\path[draw=drawColor,line width= 0.2pt,line join=round] (108.43, 19.53) --
	(108.43,146.99);

\path[draw=drawColor,line width= 0.4pt,line join=round] ( 21.74, 38.68) --
	(126.36, 38.68);

\path[draw=drawColor,line width= 0.4pt,line join=round] ( 21.74, 70.49) --
	(126.36, 70.49);

\path[draw=drawColor,line width= 0.4pt,line join=round] ( 21.74,102.30) --
	(126.36,102.30);

\path[draw=drawColor,line width= 0.4pt,line join=round] ( 21.74,134.11) --
	(126.36,134.11);

\path[draw=drawColor,line width= 0.4pt,line join=round] ( 42.61, 19.53) --
	( 42.61,146.99);

\path[draw=drawColor,line width= 0.4pt,line join=round] ( 68.94, 19.53) --
	( 68.94,146.99);

\path[draw=drawColor,line width= 0.4pt,line join=round] ( 95.27, 19.53) --
	( 95.27,146.99);

\path[draw=drawColor,line width= 0.4pt,line join=round] (121.60, 19.53) --
	(121.60,146.99);
\definecolor{drawColor}{RGB}{248,118,109}

\path[draw=drawColor,line width= 0.6pt,line join=round] ( 26.49, 48.13) --
	( 31.55, 61.52) --
	( 36.50, 67.34) --
	( 41.55, 77.01) --
	( 46.50, 82.61) --
	( 51.56, 92.47) --
	( 56.51, 98.67) --
	( 61.56, 99.94) --
	( 66.51,109.55) --
	( 71.57,113.27) --
	( 76.52,115.31) --
	( 81.58,119.89) --
	( 86.53,121.86) --
	( 91.58,123.51) --
	( 96.53,124.40) --
	(101.59,127.65) --
	(106.54,128.16) --
	(111.59,128.95) --
	(116.54,132.52) --
	(121.60,134.27);

\path[draw=drawColor,line width= 0.6pt,dash pattern=on 2pt off 2pt ,line join=round] ( 26.49, 26.24) --
	( 31.55, 36.93) --
	( 36.50, 39.89) --
	( 41.55, 39.38) --
	( 46.50, 45.39) --
	( 51.56, 48.86) --
	( 56.51, 51.94) --
	( 61.56, 54.74) --
	( 66.51, 56.78) --
	( 71.57, 59.67) --
	( 76.52, 59.90) --
	( 81.58, 62.73) --
	( 86.53, 63.59) --
	( 91.58, 65.88) --
	( 96.53, 65.91) --
	(101.59, 66.19) --
	(106.54, 65.27) --
	(111.59, 66.00) --
	(116.54, 66.00) --
	(121.60, 68.17);
\definecolor{drawColor}{RGB}{0,186,56}

\path[draw=drawColor,line width= 0.6pt,line join=round] ( 26.49, 50.26) --
	( 31.55, 64.06) --
	( 36.50, 83.21) --
	( 41.55, 89.64) --
	( 46.50, 94.28) --
	( 51.56,101.95) --
	( 56.51,107.58) --
	( 61.56,112.60) --
	( 66.51,118.84) --
	( 71.57,123.04) --
	( 76.52,123.86) --
	( 81.58,127.74) --
	( 86.53,128.48) --
	( 91.58,131.02) --
	( 96.53,132.64) --
	(101.59,133.88) --
	(106.54,136.17) --
	(111.59,139.07) --
	(116.54,140.63) --
	(121.60,141.20);

\path[draw=drawColor,line width= 0.6pt,dash pattern=on 2pt off 2pt ,line join=round] ( 26.49, 30.35) --
	( 31.55, 38.81) --
	( 36.50, 47.91) --
	( 41.55, 57.19) --
	( 46.50, 56.78) --
	( 51.56, 56.02) --
	( 56.51, 59.90) --
	( 61.56, 62.82) --
	( 66.51, 63.40) --
	( 71.57, 65.43) --
	( 76.52, 66.16) --
	( 81.58, 68.33) --
	( 86.53, 67.59) --
	( 91.58, 67.28) --
	( 96.53, 68.80) --
	(101.59, 70.43) --
	(106.54, 70.17) --
	(111.59, 70.93) --
	(116.54, 73.83) --
	(121.60, 74.31);
\definecolor{drawColor}{RGB}{97,156,255}

\path[draw=drawColor,line width= 0.6pt,line join=round] ( 26.49, 35.02) --
	( 31.55, 44.57) --
	( 36.50, 52.55) --
	( 41.55, 64.41) --
	( 46.50, 71.98) --
	( 51.56, 78.25) --
	( 56.51, 85.41) --
	( 61.56, 87.35) --
	( 66.51, 94.19) --
	( 71.57, 97.69) --
	( 76.52,101.18) --
	( 81.58,103.00) --
	( 86.53,108.40) --
	( 91.58,107.83) --
	( 96.53,111.55) --
	(101.59,119.09) --
	(106.54,120.08) --
	(111.59,123.80) --
	(116.54,121.03) --
	(121.60,121.35);

\path[draw=drawColor,line width= 0.6pt,dash pattern=on 2pt off 2pt ,line join=round] ( 26.49, 25.32) --
	( 31.55, 32.67) --
	( 36.50, 38.87) --
	( 41.55, 40.97) --
	( 46.50, 41.13) --
	( 51.56, 43.87) --
	( 56.51, 45.81) --
	( 61.56, 46.03) --
	( 66.51, 47.84) --
	( 71.57, 49.30) --
	( 76.52, 51.09) --
	( 81.58, 52.80) --
	( 86.53, 53.69) --
	( 91.58, 55.76) --
	( 96.53, 55.92) --
	(101.59, 56.59) --
	(106.54, 57.10) --
	(111.59, 57.16) --
	(116.54, 57.51) --
	(121.60, 60.25);
\definecolor{drawColor}{RGB}{0,0,0}

\path[draw=drawColor,line width= 1.1pt,dash pattern=on 4pt off 4pt ,line join=round] ( 90.00, 19.53) -- ( 90.00,146.99);
\end{scope}
\begin{scope}
\path[clip] (  0.00,  0.00) rectangle (200.66,150.49);
\definecolor{drawColor}{gray}{0.30}

\node[text=drawColor,anchor=base east,inner sep=0pt, outer sep=0pt, scale=  0.56] at ( 18.59, 36.75) {0.2};

\node[text=drawColor,anchor=base east,inner sep=0pt, outer sep=0pt, scale=  0.56] at ( 18.59, 68.56) {0.3};

\node[text=drawColor,anchor=base east,inner sep=0pt, outer sep=0pt, scale=  0.56] at ( 18.59,100.37) {0.4};

\node[text=drawColor,anchor=base east,inner sep=0pt, outer sep=0pt, scale=  0.56] at ( 18.59,132.18) {0.5};
\end{scope}
\begin{scope}
\path[clip] (  0.00,  0.00) rectangle (200.66,150.49);
\definecolor{drawColor}{gray}{0.30}

\node[text=drawColor,anchor=base,inner sep=0pt, outer sep=0pt, scale=  0.56] at ( 42.61, 12.52) {250};

\node[text=drawColor,anchor=base,inner sep=0pt, outer sep=0pt, scale=  0.56] at ( 68.94, 12.52) {500};

\node[text=drawColor,anchor=base,inner sep=0pt, outer sep=0pt, scale=  0.56] at ( 95.27, 12.52) {750};

\node[text=drawColor,anchor=base,inner sep=0pt, outer sep=0pt, scale=  0.56] at (121.60, 12.52) {1000};
\end{scope}
\begin{scope}
\path[clip] (  0.00,  0.00) rectangle (200.66,150.49);
\definecolor{drawColor}{RGB}{0,0,0}

\node[text=drawColor,anchor=base,inner sep=0pt, outer sep=0pt, scale=  0.70] at ( 74.05,  4.86) {Aquired samples from pool};
\end{scope}
\begin{scope}
\path[clip] (  0.00,  0.00) rectangle (200.66,150.49);
\definecolor{drawColor}{RGB}{0,0,0}

\node[text=drawColor,rotate= 90.00,anchor=base,inner sep=0pt, outer sep=0pt, scale=  0.70] at (  8.32, 83.26) {Validation accuracy};
\end{scope}
\begin{scope}
\path[clip] (  0.00,  0.00) rectangle (200.66,150.49);
\definecolor{drawColor}{RGB}{0,0,0}

\node[text=drawColor,anchor=base west,inner sep=0pt, outer sep=0pt, scale=  0.70] at (136.86,130.58) {Aquisition method};
\end{scope}
\begin{scope}
\path[clip] (  0.00,  0.00) rectangle (200.66,150.49);
\definecolor{drawColor}{RGB}{248,118,109}

\path[draw=drawColor,line width= 0.6pt,line join=round] (138.30,119.17) -- (149.86,119.17);
\end{scope}
\begin{scope}
\path[clip] (  0.00,  0.00) rectangle (200.66,150.49);
\definecolor{drawColor}{RGB}{0,186,56}

\path[draw=drawColor,line width= 0.6pt,line join=round] (138.30,104.71) -- (149.86,104.71);
\end{scope}
\begin{scope}
\path[clip] (  0.00,  0.00) rectangle (200.66,150.49);
\definecolor{drawColor}{RGB}{97,156,255}

\path[draw=drawColor,line width= 0.6pt,line join=round] (138.30, 90.26) -- (149.86, 90.26);
\end{scope}
\begin{scope}
\path[clip] (  0.00,  0.00) rectangle (200.66,150.49);
\definecolor{drawColor}{RGB}{0,0,0}

\node[text=drawColor,anchor=base west,inner sep=0pt, outer sep=0pt, scale=  0.56] at (154.81,117.24) {BALD};
\end{scope}
\begin{scope}
\path[clip] (  0.00,  0.00) rectangle (200.66,150.49);
\definecolor{drawColor}{RGB}{0,0,0}

\node[text=drawColor,anchor=base west,inner sep=0pt, outer sep=0pt, scale=  0.56] at (154.81,102.79) {LEA};
\end{scope}
\begin{scope}
\path[clip] (  0.00,  0.00) rectangle (200.66,150.49);
\definecolor{drawColor}{RGB}{0,0,0}

\node[text=drawColor,anchor=base west,inner sep=0pt, outer sep=0pt, scale=  0.56] at (154.81, 88.33) {max entropy};
\end{scope}
\begin{scope}
\path[clip] (  0.00,  0.00) rectangle (200.66,150.49);
\definecolor{drawColor}{RGB}{0,0,0}

\node[text=drawColor,anchor=base west,inner sep=0pt, outer sep=0pt, scale=  0.70] at (136.86, 63.53) {Model};
\end{scope}
\begin{scope}
\path[clip] (  0.00,  0.00) rectangle (200.66,150.49);
\definecolor{drawColor}{RGB}{0,0,0}

\path[draw=drawColor,line width= 0.6pt,line join=round] (138.30, 52.12) -- (149.86, 52.12);
\end{scope}
\begin{scope}
\path[clip] (  0.00,  0.00) rectangle (200.66,150.49);
\definecolor{drawColor}{RGB}{0,0,0}

\path[draw=drawColor,line width= 0.6pt,dash pattern=on 2pt off 2pt ,line join=round] (138.30, 37.67) -- (149.86, 37.67);
\end{scope}
\begin{scope}
\path[clip] (  0.00,  0.00) rectangle (200.66,150.49);
\definecolor{drawColor}{RGB}{0,0,0}

\node[text=drawColor,anchor=base west,inner sep=0pt, outer sep=0pt, scale=  0.56] at (154.81, 50.20) {conv};
\end{scope}
\begin{scope}
\path[clip] (  0.00,  0.00) rectangle (200.66,150.49);
\definecolor{drawColor}{RGB}{0,0,0}

\node[text=drawColor,anchor=base west,inner sep=0pt, outer sep=0pt, scale=  0.56] at (154.81, 35.74) {mlp};
\end{scope}
\end{tikzpicture}
\input{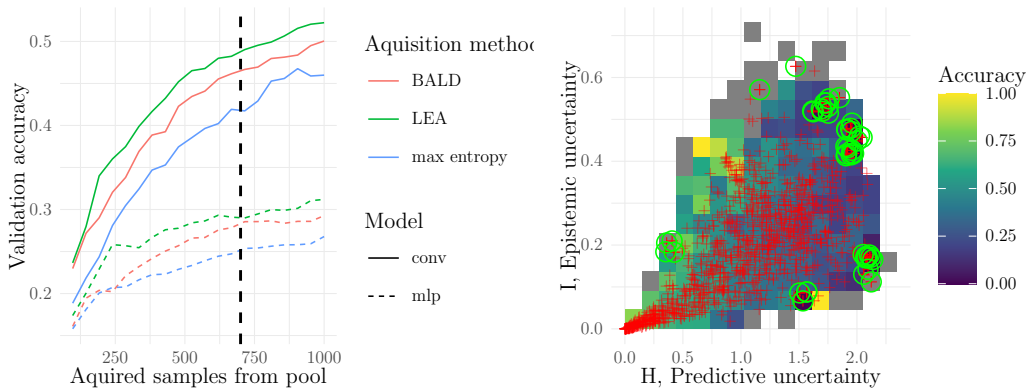}
\caption{Active learning on CIFAR10 by iterated acquisition using lowest expected accuracy (LEA), mutual information $I(x)$ (BALD) and predictive entropy $H(y| \mathcal{D}, x)$ (max entropy). Each active-learning iteration acquires
50 new samples from the pool of unlabelled data sorted by the acquisition function
calculated using the ensemble of neural networks trained on the data from the previous active-learning iteration. Our acquisition function
picks samples with lowest expected accuracy as inferred by the accuracy distribution in
the $(H, I)$-plane for the calibration dataset. The black dashed line in the left
panel corresponds to active-learning iteration 14. The right panel shows the accuracy on the calibration dataset (background tiles), pool of unlabelled data (red crosses) and acquired inputs (green circles).}
\label{fig:AL_cifar}
\end{figure}
Given the patterns in the accuracy distributions in the $(H, I)$ plane shown in Figure~\ref{fig:mnist_acc}, is it possible to use information on where the model is less
accurate to improve training?  According to the results in section \ref{sec:accuracy},
both the predictive uncertainty and the epistemic uncertainty are needed to parameterize
the accuracy. Since standard acquisition functions, such as BALD \cite{bald} and max entropy \cite{lewis1995sequential}, only
refer to the marginal uncertainty distributions, we propose here an acquisition function
that uses $H$ as well as $I$ to pick inputs from the pool of unlabelled data $D_{\text{pool}}$ corresponding to low-accuracy regions in the $(H, I)$
plane.

The goal is to increase the prediction accuracy of the model by choosing low-accuracy
inputs from $D_{\rm pool}$ to label and train on. The problem is of course that the
samples from $D_{\rm pool}$ are not labeled, therefore there is no direct way of
determining the accuracy. The idea is to create a look-up table $\mathrm{EA}(H,I)$ that parameterizes the relation
between expected accuracy and $(H, I)$. This is done using a labeled calibration dataset
$D_{\rm calibration}$, and discretising $H$ as well as $I$ to obtain the look-up table.

Using this look-up table an active-learning loop is constructed in the following way.
For each iteration, the values of $H$ and $I$ are computed for all input samples in
the unlabelled pool $D_{\rm pool}$, and the expected accuracy $\mathrm{EA}(H,I)$ is
evaluated. The input samples are then ordered by their expected accuracy, and the 50
samples with lowest expected accuracy are labelled and added to the training dataset.
A new ensemble of neural networks is trained using the updated training dataset, and a new look-up table
is created. This loop is iterated 20 times such that a total of 1000 inputs from $D_{\rm pool}$ have
been acquired.

To evaluate the proposed acquisition function LEA, we use a simple active learning setup \cite{bald} where
an ensemble of neural networks is first trained on a small class-balanced subset of 20 inputs from the CIFAR10 training
set.
For the accuracy calibration dataset, we use a random subset of 10000 samples from
the CIFAR10 training dataset, corresponding to about 17\% of the training data. We
discretise the $(H,I)$ plane into $16\times16$ regular bins limited by the minimum and
maximum values of $H$ and $I$.
As the pool of unlabelled data we use the remaining CIFAR10 training dataset after the initial inputs and accuracy calibration dataset have been removed.
We evaluate the active-learning scheme for three different acquisition functions: BALD
(maximum mutual information), max entropy (maximum predictive entropy) and our proposed
lowest expected accuracy function, LEA, described above.

Figure~\ref{fig:AL_cifar} shows the results of this procedure. The left panel shows
the evolution of the accuracy on the validation set for the dense and convolution
architectures, using the three different acquisition functions.
LEA consistently outperforms BALD, as well as active learning based solely on the maximum predictive uncertainty.
To visualize the acquisition process for a particular active-learning iteration, we
also show (right panel) the accuracy on the calibration dataset (background tiles) for
an example active-learning iteration, together with the pool of unlabelled data (red
crosses) and the inputs acquired by the acquisition function (green circles).
The panel shows that the acquired inputs are from regions in the $(H, I)$ plane that
neither maximise epistemic uncertainty nor predictive uncertainty, thus acquiring
different inputs than using BALD or max entropy as acquisition function.
This explains why LEA outperforms the other two acquisition functions: neither H nor I alone suffice to estimate the expected accuracy.

\subsection{Uncertainty}
\begin{figure}[t!]
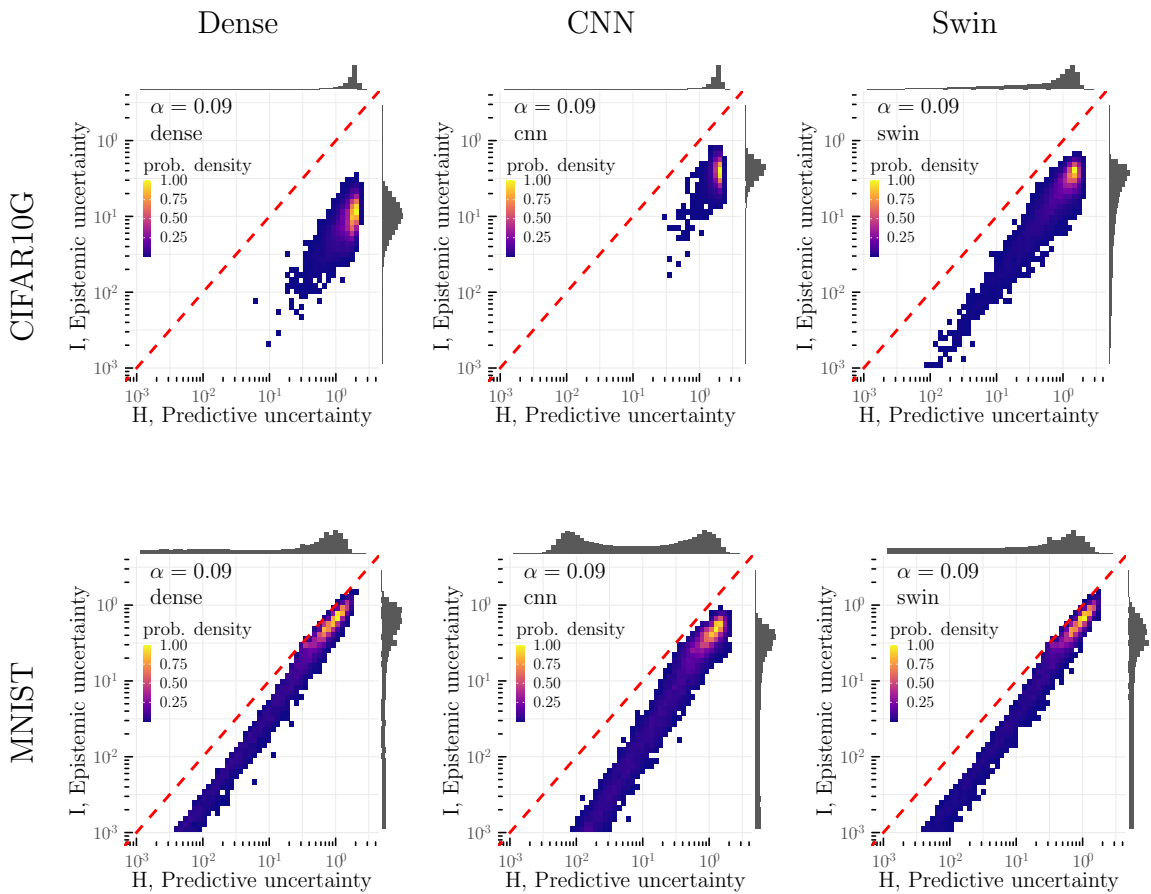

  \centering
  \begin{minipage}{\textwidth}
  \raisebox{5em}{\rotatebox{90}{CIFAR10G}}
  \begin{minipage}[b]{0.33\textwidth}
  \centering 
  Dense \\
  \input{ep/ensemble_cifar_dense_alpha_0.09.tex}
  \end{minipage}
  \hspace{-1.5em}
  \begin{minipage}[b]{0.33\textwidth}
  \centering 
  CNN \\
  \input{ep/ensemble_cifar_cnn_alpha_0.09.tex}
  \end{minipage}
  \hspace{-1.5em}
  \begin{minipage}[b]{0.33\textwidth}
  \centering 
  Swin \\
  \input{ep/ensemble_cifar_swin_alpha_0.09.tex}
  \end{minipage}
  \end{minipage}
\begin{minipage}[t]{\textwidth}
\raisebox{5em}{\rotatebox{90}{MNIST}}
  \input{ep/ensemble_dense_alpha_0.09.tex}
  \hspace{-1.5em}
  \input{ep/ensemble_cnn_alpha_0.09.tex}
  \hspace{-1.5em}
  \input{ep/ensemble_swin_alpha_0.09.tex}
  \end{minipage}
  
\caption{Joint distribution $(H, I)$, corresponding to predictive uncertainty and
epistemic uncertainty, for different neural networks trained on CIFAR10G (top row), MNIST (bottom row) and evaluated
on noised data with noise parameter $\alpha=0.09$. The posterior distributions are
approximated using ensembling. Inset in each frame is the noise level ($\alpha$).
The dashed diagonal indicates the bound on epistemic uncertainty
from Eq.~\eqref{eq:bound}. The marginal probability density distributions are shown above and to the right of each panel. \label{fig:method_cifar}\label{fig:method_mnist}}
\end{figure}

To understand how the uncertainty distribution in Figure~\ref{fig:cifar_acc}
correlates with noise level, we now turn to the uncertainty density distributions
at a fixed noise level. Figure~\ref{fig:method_cifar} shows, for different neural-
network architectures, the joint distribution of the predictive uncertainty defined
by Eq.~\eqref{eq:posterior_entropy_approx} and epistemic uncertainty as defined by
Eq.~\eqref{deltaH} in the $(H, I)$ plane. The uncertainty measures are calculated using
approximations to the Bayesian posterior provided by the ensemble method as explained
in section \ref{sec:uncertainty_quantification}. The joint distributions in Figure~\ref{fig:method_cifar} are evaluated on the test set of CIFAR10G shifted by impulse
noise with noise parameter $\alpha=0.09$, corresponding to $9\%$ of pixels set to the
maximum grayscale intensity, see Figure~\ref{fig:mnist_salt}. The joint distributions in
Figure~\ref{fig:method_cifar} allow us to separate the origin of uncertainty: a sample
with large value of $I$ on the vertical axis has higher epistemic uncertainty, whereas
a sample on the horizontal axis has no epistemic uncertainty. The dashed red diagonal
line indicates the bound on epistemic uncertainty imposed by Eq.~\eqref{eq:bound}. In
particular this means that a sample close to the epistemic bound is dominated by epistemic
uncertainty.

The first panel in the top row of Figure~\ref{fig:method_cifar} shows the joint distribution for the dense model. The joint distribution is shifted towards higher predictive uncertainty and lower epistemic uncertainty. The top-center panel in Figure~\ref{fig:method_cifar} shows the joint uncertainty distribution for the convolutional neural network. Here, the perceived epistemic uncertainty is larger compared to the other models.
The right-most panel in Figure~\ref{fig:method_cifar} shows that the attention-based model exhibits similar epistemic uncertainty to the convolutional neural network in the center panel. The attention-based model does however perceive a lower aleatoric uncertainty, although this is not accompanied by a significant increase in accuracy.

Turning to MNIST, Figure~\ref{fig:method_mnist} shows the corresponding joint distributions for the three architectures at noise level $\alpha=0.09$.
The first panel shows joint distributions of predictive and epistemic uncertainty for the dense model. In contrast to the dense model on CIFAR10G, the epistemic uncertainty is larger than the other models and the predictive uncertainty is smaller. The center panel shows the joint distribution for the convolutional neural network, and the right-most panel the attention-based Swin model.

To evaluate the dependence of the uncertainty quantification on the training dataset we calculate the induced shift in joint distribution of predictive and epistemic uncertainty when we apply the same data-distributional shift to neural networks trained on MNIST and CIFAR10G.
The shift from $\alpha=0$ to $\alpha=0.09$ in mean predictive and epistemic uncertainty
is quantified in Table~\ref{tab:alpha_comparison}. For the dense and Swin model we
observe a significantly smaller shift in both $H$ and $I$ on CIFAR10G compared to MNIST,
whereas the convolutional neural network perceives a larger shift on CIFAR10G compared
to MNIST. 

The dense and Swin model both contain more parameters for CIFAR10G compared to
MNIST stemming from the 4 extra pixels in both spatial directions. One possible method
to keep the parameter count constant would be to resample the resolution of the images,
however this would also introduce some dependence on the resampling method. Here we
keep the original resolution for simplicity and leave the possibility of resampling for
future work.

\begin{table}[t!]
  \centering
  \begin{tabular}{|ll||lll|}
    \hline
    Architecture & Dataset & \multicolumn{1}{c}{$\alpha$} & \multicolumn{1}{c}{$\Delta \bar{H}$} & \multicolumn{1}{c|}{$\Delta \bar{I}$} \\
    \hline
    Dense & MNIST & $0.09$ & $0.49$ & $0.33$ \\
Dense & CIFAR & $0.09$ & $0.08$ & $0.03$ \\
    \hline
    CNN & MNIST & $0.09$ & $0.65$ & $0.24$ \\
CNN & CIFAR & $0.09$ & $0.85$ & $0.25$ \\
    \hline
    Swin & MNIST & $0.09$ & $0.63$ & $0.41$ \\
Swin & CIFAR & $0.09$ & $0.22$ & $0.20$ \\
    \hline
  \end{tabular}
  \caption{Change in mean predictive and epistemic uncertainty, $\Delta \bar{H}$ and $\Delta \bar{I}$, for fully connected (Dense), convolutional (CNN) and attention-based (Swin) models from $\alpha=0$ to $\alpha=0.09$. Posterior approximation by ensembling.\label{tab:alpha_comparison}}
\end{table}
\

\section{Discussion}

\subsection{Accuracy}
Our results allow us to draw three conclusions regarding the connection between the joint distribution of epistemic and predictive uncertainty, and prediction accuracy.

First, looking at the joint distribution helps to identify samples where a given model is more accurate. One important goal of uncertainty quantification is to assess when the predictions of a model can be trusted. It is natural to expect that a prediction is more accurate when the predictive uncertainty is low. We show that the joint distribution of epistemic and predictive uncertainty can identify accurate predictions while predictive uncertainty or epistemic uncertainty alone cannot. This can be seen in Figure~\ref{fig:cifar_acc}, where the joint distribution clearly resolves where the models are accurate, and where they are not.  Figure~\ref{fig:cifar_acc} show that the projection of the joint distribution on either the axis of predictive uncertainty or the axis of epistemic uncertainty mixes samples with high and low accuracy. As a consequence, selecting for predictive uncertainty or epistemic uncertainty alone, is not as effective in identifying where a given model is more accurate. This explains the findings in e.g. Refs. \citep{ghoshal2, krishnan2020improving}, regarding the clustering of incorrect predictions at high uncertainty, and also explains why it is difficult to achieve higher precision using the marginalized measures of epistemic and predictive uncertainty, as observed in Ref.~\cite{nair2020exploring}: samples with small values for the marginalized distributions can still have accurate model predictions.

Second, for the attention-based model it is difficult to find a threshold of the projected distribution on either predictive uncertainty or epistemic uncertainty that results in high accuracy. The top-right panels in Figure~\ref{fig:cifar_acc} show a distinct structure for the accuracy conditioned on the joint distribution where the marginalized distribution on either axis mixes samples of high and low accuracy. This has implications for active learning. In the present example, choosing samples based on e.g.\ posterior predictive uncertainty alone is inefficient, because it results in training on data regions where the model is already accurate.

Third, by training on data regions where the accuracy is low, there is potential to increase overall prediction accuracy. The joint distribution of epistemic and predictive uncertainty can be used to identify such data regions where the model accuracy can be improved, also for data that is not yet labeled. Figure~\ref{fig:cifar_acc} show that the regions are specific to each network architecture, and to the dataset used for training. The mixing of high and low accuracy prediction when projecting the distributions in Figure~\ref{fig:cifar_acc} also explain the difficulty in achieving better sample efficiency by using the decomposition into aleatoric and epistemic uncertainty observed in Ref.~\cite{beluch2018power}: The projected distributions will contain regions of high prediction accuracy for small uncertainties. An active learning scheme that makes use of uncertainty quantification could therefore benefit from selection using the joint distribution.

\subsection{Active learning}
Since prediction accuracy depends on both epistemic and predictive uncertainty, we proposed an acquisition function for active learning that uses a parameterization of expected accuracy in terms of both conditional mutual information and predictive entropy.
Our proposed LEA (lowest expected accuracy) acquisition function, defined in
section \ref{sec:method_AL}, outperforms acquisition functions using marginal uncertainty distributions.
The left panel in Figure~\ref{fig:AL_cifar} shows that our calibrated accuracy acquisition produce samples
that enable the neural networks achieve higher validation accuracy with less training
samples. The right panel in Figure~\ref{fig:AL_cifar} shows that LEA picks inputs from
regions that would have been missed by BALD (acquired inputs do not maximise mutual
information) and max entropy (acquired inputs do not maximise predictive entropy). We
stress that the proposed acquisition function only considers single sample uncertainty,
we expect that e.g accuracy gain per iteration can be further improved by incorporating expected accuracy in state of the art acquisition
functions such as BatchBALD \cite{batchbald} or EPIG \cite{smith2023prediction}. Also
note that the comparison with BALD and max entropy in Figure~\ref{fig:AL_cifar} can
be argued to be unfair in the sense that LEA uses information about the targets from
the calibration dataset.  This could be amended by calibrating on the current training
dataset instead of a held-out calibration dataset.

\subsection{Uncertainty}
We can draw two conclusions from our results regarding how uncertainty quantification depends on model architecture.

First, the origin of uncertainty is not objective. Table~\ref{tab:method_cifar_validation} shows that there are significant differences in the perceived origin of uncertainty between the different model architectures for in-domain data. For CIFAR10G, the fully connected neural network perceives a higher degree of aleatoric uncertainty, and low epistemic uncertainty compared to the convolutional and attention-based models. This shows that the origin of uncertainty depends on the model architecture: the fully connected neural network struggles to express the relationship between inputs and classes of CIFAR10G, and thus perceives a higher degree of aleatoric uncertainty. Even though the model has low accuracy, the ensemble posterior approximation show a low epistemic uncertainty. Thus we do not expect to be able to salvage the accuracy of the fully connected neural network by increasing the size of the dataset.
For CIFAR10G, the top row of Figure~\ref{fig:method_cifar} and Table~\ref{tab:method_cifar} show that for a moderate data-distributional shift, the fully connected neural network perceives a higher degree of aleatoric uncertainty and low epistemic uncertainty compared to the convolutional and attention-based models. Here the dense architecture is already perceiving the validation data as aleatoric noise and thus continues to do so for further shifts.

Second, the different model posteriors agree about the origin of uncertainty when the dataset complexity is low. In the case of MNIST, the bottom row of Figure~\ref{fig:method_mnist} and Table~\ref{tab:method_mnist} show that all neural-network architectures considered here have similar joint distributions of predictive and epistemic uncertainty. In other words, the MNIST data has relatively low complexity, as evident by the higher average accuracies in Table~\ref{tab:architectures}, and therefore all three architectures succeed in capturing the important features. As a consequence, all three architectures agree about the origin of uncertainty.

In addition, there are two conclusions we can draw from our results regarding how predictive and epistemic uncertainty depend on the dataset. First, the relative change in perceived uncertainty under a fixed data-distributional shift depends on the dataset, and varies with model architecture. Comparing the change in the joint distribution under a fixed data-distributional shift in Table~\ref{tab:alpha_comparison}, we observe that a given model architecture understands the same data-distributional shift in different ways, depending on the underlying training dataset. There is an asymmetry in this shift sensitivity: only the convolutional model shows a stronger sensitivity of the predictive and epistemic uncertainty when evaluated on CIFAR10G compared to MNIST. Furthermore, the fully connected model perceives the impulse noise on CIFAR10G very differently from impulse noise on MNIST.
The distributional shift for MNIST induces a large change in both epistemic and predictive uncertainty, whereas for CIFAR10G the induced shift in uncertainty is significantly smaller. This can be explained by CIFAR10G containing realistic digital images, hence we expect impulse noise to be more in-domain than for MNIST. The converse is true for the convolutional model, where the induced shift is larger for CIFAR10G. In summary, the sensitivity to a particular data-distributional shift depends on the data domain and on neural-network architecture.
Even though this difference in perceived relative change of uncertainty is present for widely different data, it also has implications for different domains in the same training dataset, something that would be interesting to quantify in more detail.

Second, robustness of prediction accuracy under data-distributional shifts for a given model depends on the dataset. For CIFAR10G, the higher accuracy of the fully connected neural network on noised data in Table~\ref{tab:method_cifar} shows that this architecture is more robust against this particular distributional shift, even though the model is less accurate close to the training domain as seen in Table~\ref{tab:method_cifar_validation}. For MNIST, the convolutional model is instead more robust than both the dense and attention-based architectures as seen in Table~\ref{tab:method_mnist}.

\section{Conclusions}

Posterior predictive entropy and mutual information are used extensively as measures of predictive uncertainty and epistemic uncertainty to assess the uncertainty and performance of neural networks and their predictions \citep{ABDAR2021243}.
We introduced the joint distribution of predictive uncertainty and epistemic uncertainty and demonstrated how it is related to model accuracy.

Previous work have shown that in both active learning \citep{chitta2018large,pmlr-v80-depeweg18a} and predictive uncertainty estimation \citep{ghoshal,ghoshal2}, it is
difficult to formulate a general strategy making use of a decomposition of uncertainty
into predictive and epistemic parts. Our results explain why it is difficult to use
predictive or epistemic uncertainty separately as a measure of model efficacy by
showing how the joint distribution resolves information that is lost in projections.
We showed that the joint distribution contains information about when a neural network
is accurate, and that the distribution is specific to the particular choice of neural-
network architecture and dataset.

To test these insight, we proposed a novel acquisition function using expected accuracy
parameterized in terms of epistemic and predictive uncertainty. The proposed acquisition
function outperforms two common acquisition functions based on marginal uncertainty
distributions.

In addition, we also demonstrated that the origin of uncertainty is not objective:
different model architectures will disagree about the aleatoric and epistemic
uncertainty. Furthermore, for a given model, the sensitivity of the uncertainty
quantification to a specific type of data-distributional shift depends on the underlying
training dataset.

We conclude by mentioning the most important open questions. First, is it possible to explain how uncertainty quantification depends on architecture from the mathematical theory of neural networks, and how to use this to build architectures that target robust uncertainty quantification. Second, given the recent rise of attention-based architectures across multiple domains such as natural language processing and computer vision, it is of utmost importance to properly understand their posteriors and related uncertainty measures.
Third, in practical application, uncertainty quantification using the Bayesian posterior depends on accurate posterior approximations. Here we used ensembles as a robust baseline, but there is a growing need for accurate and computationally effective posterior approximation methods.  Finally, entropy and mutual information are two measures of uncertainty derived from the high-dimensional model posterior and the posterior predictive distribution. Whether there might be other, complementary or more informative, derived quantities that can capture the uncertainty of artificial neural networks better also remains an interesting question.

In summary, the joint distribution of predictive and epistemic uncertainty can inform on neural-network efficacy when calibrated for a given dataset and architecture.

\

\section*{Acknowledgments}
BM and HL were supported in part by Vetenskapsrådet (grant numbers 2017-3865 and 2021-4452). OB was supported by Vetenskapsrådet (grant number 2017-05162).
HL would like to thank Erik Werner for enlightening discussions.
\clearpage
\appendix

\section{Ensemble posterior}
\label{sec:ensemble}
The Bayesian mean in Eq.~\eqref{eq:bayesian_mean} can be approximated by Monte-Carlo
sampling of the integration over model parameters. Given $N$ parameter samples $
\theta^{(i)}$ of equal posterior probability, the Bayesian mean can be approximated as
in Eq.~\eqref{eq:ppd}.
\begin{align}
  p(y| x, \mathcal{D}) & = \int_{\theta} f(y, x; \theta) p(\theta | \nobracket \mathcal{D}) \tmop{d\theta} \approx \frac{1}{N} \sum_{i = 0}^N f(y, x; \theta^{(i)}) \label{eq:ppd}
\end{align}

A frequently employed method to sample from the model space is to train an ensemble
{\citep{lakshminarayanan2017simple,hoffmann2021deep}} of $N$ identical neural networks using different
initial parameter values $\theta_{\tmop{initial}}^{(i)}$ and regularized by quadratic loss on parameter norm corresponding to the Gaussian prior. Training these
neural networks to maximize the likelihood of the training data gives a set of
parameters $\theta^{(i)}$, that then provides an approximation to the Bayesian mean
by Eq.~\eqref{eq:ensemble}, where $\theta^{(i)}$ are the ensemble member parameters.

\begin{equation}
  p(y=c|x, \mathcal{D}) \approx \frac{1}{N} \sum_{i = 0}^N f_{c}(x; \theta^{(i)}). \label{eq:ensemble}
\end{equation}

Here it is assumed that the likelihood of the minima attained by $\theta^{(i)}$ are
equally probable. Note that it is not clear a priori whether the minima $\theta^{(i)}$ are
degenerate, but for the regime of neural networks for visual perception
it is typically the case that they are not \citep{loss_modality}.

\section{Toy model posterior}\label{sec:toy_model}
The relation between mutual information of the parameter posterior and mutual information of the posterior predictive in Eq.~\eqref{deltaH} provides a way of calculating the expected change in entropy of the high-dimensional parameter posterior $p(\theta | \mathcal{D})$ in terms of the typically lower-dimensional posterior predictive distribution $p(y|x,\mathcal{D})$ and the likelihood $p(y|x, \theta)$.
To illuminate this relation, and evaluate the involved quantities in closed-form, we present a detailed verification for a simple toy model. This also serves as an illustration of the computational complexity involved in computing the Bayesian posterior directly.

\begin{figure}[t]
  \centering
  \includegraphics{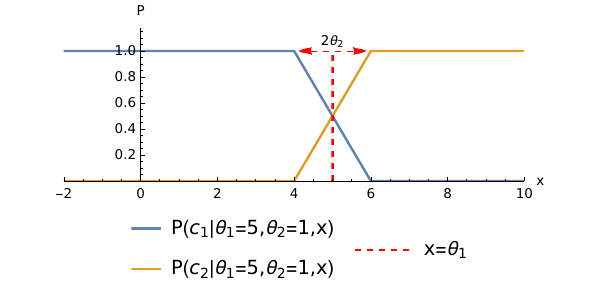}
  \caption{Example of the likelihood $p(c|\theta_{1}, \theta_{2}, x)$ in equations~\ref{eq:l1} and~\ref{eq:l2} over the two classes $c\in\{c_{1},c_{2}\}$ for different values of $x$ with model parameters $\theta_{1}=5$ and $\theta_{2}=1$ corresponding to the decision boundary.\label{fig:toy_model}}
\end{figure}

The toy problem consists of classifying points on the real line into two classes ${c_{1},c_{2}}$ and we use a simple two-parameter linear model
\begin{align}
p\left(\left.c_1\right|x,\theta _1,\theta _2\right)&=
\begin{cases}
 1 & x-\theta _1<-\theta _2, \\
 \frac{\left(x-\theta _1+\theta _2\right)}{2\theta _2} & \left|x-\theta _1\right|<\theta _2, \\
 0 & x-\theta _1>\theta _2, \label{eq:l1}\\
\end{cases} \\
p\left(\left.c_2\right|x,\theta _1,\theta _2\right) &=1.0 - p\left(\left.c_1\right|x,\theta _1,\theta _2\right). \label{eq:l2}
\end{align}
By construction, this model has a strong prior for samples of  class 1 being located to the left of a decision boundary at $\theta_{1}$ and class 2 to the right. See Figure~\ref{fig:toy_model} for the resulting probability distributions for a particular choice of the model parameters $\theta_{1}$ and $\theta_{2}$.

\begin{figure}[t]
  \centering
  \includegraphics{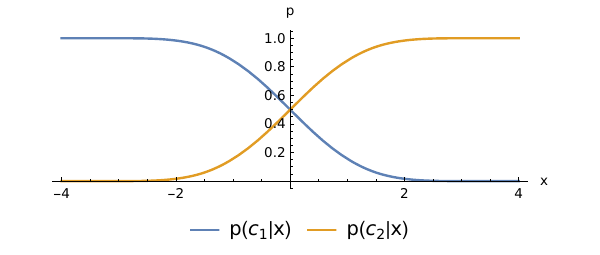}
  \caption{Prior class distribution given the uniform prior $p(\theta_{1},\theta_{2})$ on the model parameters.\label{fig:class_prior}}
\end{figure}

Let $\theta_{1}$ be uniformly distributed on $[-1, 1]$ and $\theta_{2}$ on $[\frac{1}{2}, 2]$. With this prior on the parameters, we can calculate the prior predictive distribution over the two classes by
\begin{equation}
  p(\left.c\right|x) = \int_{\theta_{1}}\int_{\theta_{2}}p(c|x, \theta_{1},\theta_{2}) p(\theta_{1},\theta_{2})\mathrm{d}\theta_{1}\mathrm{d}\theta_{2},
\end{equation}
visualized in Figure~\ref{fig:class_prior}, where we see that the prior parameter distribution results in a smooth prior predictive distribution.

Suppose we observe class $c_{1}$ at $x_{1} = 2$, we can then calculate a posterior distribution for the model parameters

\begin{equation}
  p\left(\theta _1,\theta _2|\left\{x_1,c_1\right\}\right)= \frac{p\left(c_1|x_1,\theta _1,\theta _2\right)p\left(\theta _1,\theta _2\right)}{p\left(c_1|x_{1}\right)}
\end{equation}
where we have used Bayes theorem to express the parameter posterior in terms of conditional probabilities that can be calculated explicitly.

\begin{figure}[t]
  \centering
  \includegraphics{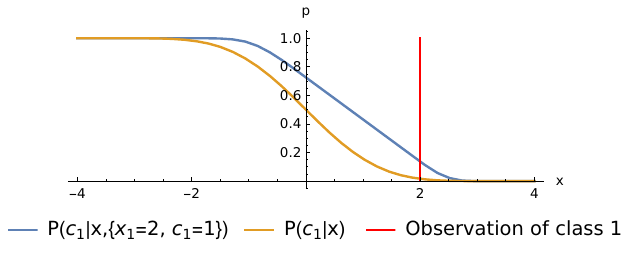}
  \caption{Posterior predictive distribution given one observation of class $c_{1}$ at $x_{1} = 2$ compared to the prior predictive distribution. The observation of class 1 to the right shifts the posterior in this direction.\label{fig:posterior_output}}
\end{figure}

With this posterior we calculate the posterior predictive distribution in Eq.~\eqref{eq:bayesian_mean} of the introduction, resulting in a slightly shifted distribution for class 1 in Figure~\ref{fig:posterior_output}, compared to the class prior in Figure~\ref{fig:class_prior}.

Using the toy model we can now explicitly verify the relation between Eq.~\eqref{eq:delta_H_theta} and~\eqref{deltaH}. The expected entropy difference in Eq.~\eqref{eq:delta_H_theta} is given by
\begin{multline}
  I(\theta_{1},\theta_{2}|x) = \int_{\theta_{1}}\int_{\theta_{2}}p(\theta_{1},\theta_{2})\log(p(\theta_{1},\theta_{2}))\mathrm{d}\theta_{1}\mathrm{d}\theta_{2}\\ -\sum_{i=1,2}p(c_{i}|x)\int_{\theta_{1}}\int_{\theta_{2}} p\left(\theta_{1},\theta_{2}|\left\{x_1,c_i\right\}\right) \log\left(p\left(\theta_{1},\theta_{2}|\left\{x_1,c_i\right\}\right)\right) \,\mathrm{d}\theta_{1}\mathrm{d}\theta_{2} \label{eq:delta_H_theta_toy}
\end{multline}
and the posterior predictive entropy difference in Eq.~\eqref{deltaH} is given by
\begin{multline}
  I(c|x) = \sum_{i=1,2} p(c_{i}|x) \log(p(c_{i}|x))\\ - \int_{\theta_{1}}\int_{\theta_{2}} \left(\sum_{i=1,2} p(c_{i}|x, \theta_{1},\theta_{2}) \log(p(c_{i}|x, \theta_{1},\theta_{2})) \right)p(\theta_{1}, \theta_{2})\,\mathrm{d}\theta_{1}\mathrm{d}\theta_{2}\label{eq:delta_H_c_toy}
\end{multline}
Note that in this case, where we compute the entropy difference when adding a single observation Eq.~\eqref{eq:delta_H_c_toy} only uses the prior distribution.

\begin{figure}[t]
  \centering
  \includegraphics{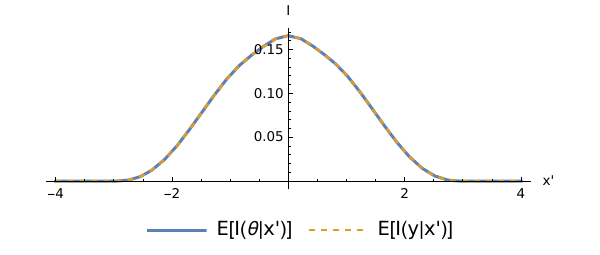}
  \caption{Entropy difference $I(\theta_{1},\theta_{2}|x)$ and $I(c|x)$ for the toy model quantifying the epistemic uncertainty for different $x$ using only the prior. \label{fig:delta_H_one}}
\end{figure}

Numerically evaluating these expressions gives Figure~\ref{fig:delta_H_one} where the two curves are indistinguishable, as expected.

Figure~\ref{fig:delta_H_one} shows that the epistemic uncertainty is largest close to the decision boundary of the prior. This can be understood intuitively by the fact that the model and parameter priors are such that adding observations of class 1 far to the left (or class 2 far to the right) does not add new information.

\begin{figure}[t]
  \centering
  \includegraphics{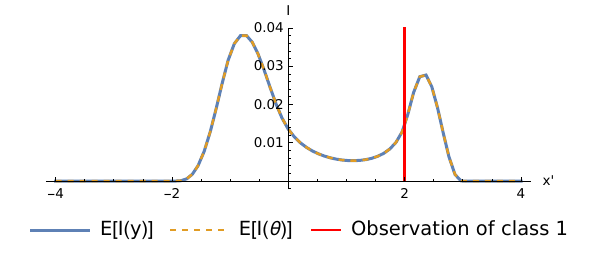}
  \caption{Entropy difference $I(\theta_{1},\theta_{2}|x, \{x_{1}=2, c=1\})$ and $I(c|x, \{x_{1}=2, c=1\})$ for the toy model quantifying the epistemic uncertainty for different $x$ after a single observation $\{x=2, c=1\}$. \label{fig:delta_H_two}}
\end{figure}

Continuing, we can perform the same calculation but instead add a new observation on top of the first one. Assuming independent observations the posterior now becomes

\begin{equation}
p\left(\theta _1,\theta _2|\left(x_1,c_1\right),\left(x_2, c_2\right)\right)= \frac{p\left(c_1|x_1,\theta _1,\theta _2\right)p\left(c_2|x_2,\theta
    _1,\theta _2\right)p\left(\theta _1,\theta _2\right)}{p\left(\left(x_1,c_1\right),\left(x_2, c_2\right)\right)}
\end{equation}

and carefully calculating the entropy differences now instead results in the epistemic uncertainty in Figure~\ref{fig:delta_H_two}. Note first that the two expressions are still in excellent agreement. The observation of class 1 at $x=2$ is in tension with the prior which can be seen by the bi-modal epistemic uncertainty.

\begin{figure}[t]
  \centering
  \includegraphics{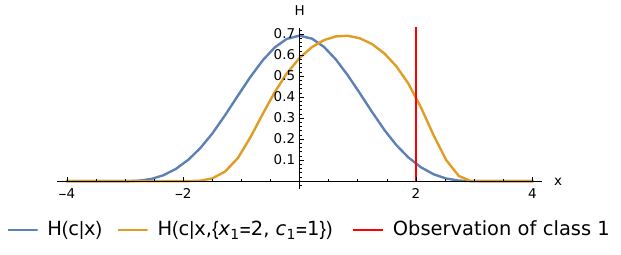}
  \caption{Entropy of the posterior predictive distribution using the prior $H(c|x)$ and after one observation $H(c|x, \{x_{1}=2, c=1\})$ for the toy model quantifying aleatoric and epistemic uncertainty for different $x$. \label{fig:entropy_output_toy}}
\end{figure}

The epistemic uncertainty can be compared to the entropy of the posterior predictive distribution in Figure~\ref{fig:entropy_output_toy} which peaks in the region between the prior decision boundary and the observed class 1 sample.

\

{
  \small

}

\end{document}